\documentclass{article}
\pdfoutput=1  

\usepackage{graphicx}
\usepackage{cite}
\usepackage{amsmath}

\newcommand{\widefigwidth}{5.0in}
\newcommand{\figwidth}{3.5in}

\newcommand{\eq}[1]{Eq.~(\ref{eq.#1})} 
\newcommand{\fig}[1]{Fig.~\ref{fig.#1}}

\newcommand{\tbl}[1]{Table~\ref{table.#1}}

\newcommand{\sect}[1]{Section~\ref{sect.#1}}
\newcommand{\sectA}[1]{Appendix~\ref{sect.#1}}
\newcommand{\sectlabel}[1]{\label{sect.#1}}
\newcommand{\eqlabel}[1]{\label{eq.#1}}
\newcommand{\figlabel}[1]{\label{fig.#1}}
\newcommand{\tbllabel}[1]{\label{table.#1}}

\renewcommand\vec{\mathbf}
\newcommand{\norm}[1]{\left\lVert#1\right\rVert}

\newcommand{\density}{\ensuremath{\rho}}
\newcommand{\viscosity}{\ensuremath{\eta}}
\newcommand{\viscosityKinematic}{\ensuremath{\nu}}  

\newcommand\Rey{\mbox{Re}}  

\newcommand\relPos{\textnormal{rp}}

\newcommand{\corr}{\textnormal{cor}}
\newcommand{\lc}{\textnormal{lc}} 
\newcommand{\Pbranch}{\ensuremath{P_{\textnormal{branch}}}}

\newcommand{\meter}{\mbox{m}}
\newcommand{\micron}{\mbox{$\mu$m}}
\newcommand{\second}{\mbox{s}}
\newcommand{\millisecond}{\mbox{ms}}
\newcommand{\kilogram}{\mbox{kg}}
\newcommand{\pascal}{\mbox{Pa}}

\newcommand{\radian}{\mbox{rad}}

\title{Identifying Vessel Branching from Fluid Stresses on Microscopic Robots}
\author{Tad Hogg\\
{\small Institute for Molecular Manufacturing}\\{\small Palo Alto, CA}
}

\begin{document}
\maketitle

\begin{abstract}

Objects moving in fluids experience patterns of stress on their surfaces determined by the geometry of nearby boundaries. Flows at low Reynolds number, as occur in microscopic vessels such as capillaries in biological tissues, have relatively simple relations between stresses and nearby vessel geometry. Using these relations, this paper shows how a microscopic robot moving with such flows can use changes in stress on its surface to identify when it encounters vessel branches.

\end{abstract}

\section{Introduction}

Microscopic devices small enough to pass through the circulatory system are useful for biological research and medicine~\cite{freitas99,martel07,monroe09,hill08,schulz09}.
For instance, nanoparticles can precisely deliver drugs~\cite{allen04}. 
More elaborate applications could arise from devices with a full range of robotic capabilities, including sensing, computation, communication and locomotion. 
These microscopic robots could sense a variety of signals, including chemicals on cells~\cite{park08}, fluid shear~\cite{korin12}, light or temperature~\cite{sershen00}.
Some existing devices~\cite{koman18,jager00,martel07a,sitti15} demonstrate these capabilities in small volumes, though they are generally too large to fit through capillaries.
Smaller demonstrated devices~\cite{andrews18,ke18,li18,martel14,thubagere17} have more limited sensing and computational capabilities than their larger counterparts.
Building on these demonstrations, proposed future microscopic robots could provide a wide range of capabilities in a volume small enough to pass through capillaries~\cite{freitas99}.

Applications of microscopic robots could improve their precision if robots can identify type of tissue they are passing through. For circulating robots in the vasculature, especially useful identification methods are those available to the robot from properties measurable from within the vessels. 
For instance, the geometry of capillaries differs among organs~\cite{augustin17} and between normal and tumor vasculature~\cite{nagy09,jain14}. 
Thus robots able to determine vessel geometry could supplement other available information, such as chemicals, to more accurately identify different types of tissue.

An important aspect of vessel geometry is when a vessel splits into branches, or when several vessels merge into a larger one.
Microscopic robots might detect branches acoustically~\cite{freitas99}, though interpreting reflected signals could require significant computation in the presence of multiple reflections, scattering and the small difference in acoustic impedance between the fluid and walls of tiny vessels.
As a complementary approach, this paper describes and evaluates how robots could detect branches from changes in the patterns of fluid-induced stresses on their surfaces.

Fluid stresses provide useful information at larger scales, e.g., for fish~\cite{bleckmann09} and underwater vehicles~\cite{yang06a,vollmayr14}.  Micromachine sensors motivated by those of fish can detect relatively small changes in fluid motion~\cite{kottapalli14}.
These fluid stresses provide information about the environment within about a body length of the object in the fluid~\cite{sichert09}, although extracting this information requires significant computation~\cite{bouffanais10}.
Another example of sensing fluid motion is the use of artificial whiskers, modeled on the geometry of seal whiskers, to estimate the direction, size and speed of moving objects from their wakes~\cite{eberhardt16}.

For microscopic objects in fluids, viscous rather than inertial forces dominate the flow~\cite{dusenbery09,purcell77}. Viscous flows have a linear relation between stresses and velocities~\cite{happel83}, allowing simpler computations to interpret stresses than required for larger-scale flows. This simplicity enables stress-based navigation by microscopic robots~\cite{hogg18} in spite of their computational limits, both in terms of operating speed and memory.
Moreover, viscous flow generally has gentle gradients, so the effect of boundaries extends relatively far into the fluid. This potentially allows a microscopic robot to use stresses to infer properties of the boundaries at larger distances, relative to its size, than is feasible for larger robots.

In the remainder of this paper, the next two sections describe typical geometric parameters of tiny vessels and the stresses on the surface of an object moving with the fluid in such vessels. The following two sections show how a robot can use stress measurements to identify when it is passing vessel branches. \sect{accuracy} evaluates the performance of this approach. The final section discusses possible extensions to more complicated scenarios.
The appendix provides details on the scenarios used for training and testing, and of the classifier used to identify branches.

\section{The Geometry of Microscopic Vessels}
\sectlabel{geometry}

This paper examines spherical robots in moving fluids similar to water, with parameters given in \tbl{fluid parameters}. 
We numerically evaluate robot behavior in vessels with geometry comparable to that of short segments of capillaries. These vessels generally have radii of curvature of tens of microns~\cite{pawlik81}, and when they split, they typically split into just two branches~\cite{cassot06}.
The  branches have a larger total cross section than the main vessel~\cite{murray26}, leading to slower flows in the branches~\cite{sochi15}.

For simplicity, we focus on planar vessel geometry. Specifically, incoming and outgoing axes of curved vessels are in the same plane. Similarly, branching vessels have the axes of the main vessel and the branches in the same plane.

\begin{figure}
\centering
\includegraphics[width=\widefigwidth]{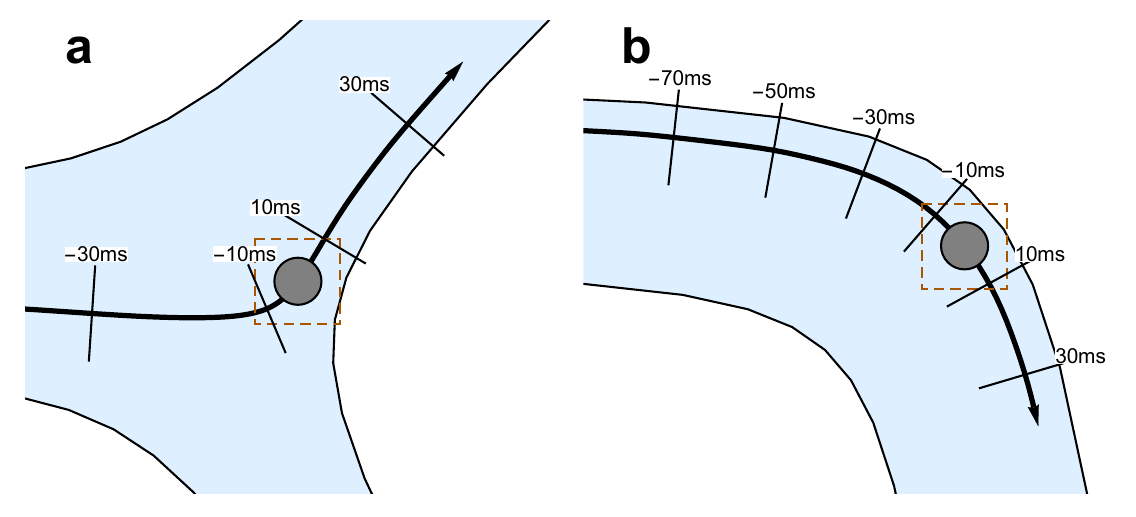}
\caption{Robot motion through (a) branched and (b) curved vessels.
The curved arrow shows a portion of the path of the robot's center as it moves through the vessel with the fluid.
The ticks along the paths indicate times, in milliseconds, before or after the location of the robot indicated by the gray disk.
The dashed rectangles indicate the parts of the vessel shown in \fig{flow}. The robot has radius $1\,\micron$ and the vessel inlets, to the left of the sections shown in the figure, both have diameters of $7.8\,\micron$.
}\figlabel{vessel geometry}
\end{figure}

\fig{vessel geometry} shows examples of the vessel geometries considered here.
The fluid flow speed for these two cases is chosen so the robots have the same average stress magnitudes on their surfaces at the indicated position along each paths. Specifically, the maximum speed at the inlet is $1000\,\micron/\second$ and $530\,\micron/\second$, for the branch and curve, respectively.
At the position of the robot shown in the branch, the robot moves at $189\,\micron/\second$ and rotates with angular velocity $-34\,\radian/\second$, i.e., clockwise.
For the curve, these values are  $186\,\micron/\second$ and $+39\,\radian/\second$.

For developing and testing a branch classifier based on stresses on the robot's surface, we create samples of robots in branch and curve vessels. \sectA{samples} describes the parameters for the vessel geometry, initial robot position and fluid flow speed used to create these samples. These vessels are similar in size to the examples of \fig{vessel geometry}.

\begin{table}
\centering
\begin{tabular}{lcc}
density		& $\density$	&$10^3 \,\kilogram/\meter^3$	\\
viscosity		& $\viscosity$	&$10^{-3}\,\pascal \cdot \second$	\\
kinematic viscosity		& $\viscosityKinematic =\viscosity/\density$ & $10^{-6}\,\meter^2/\second$ \\
vessel diameter	&$d$		& $5\text{--}10\,\micron$\\
maximum flow speed	& $u$	& $200\text{--}2000\,\micron/\second$\\
Reynolds number	&$\Rey=u d/\viscosityKinematic$	& $<0.04$\\ 
robot radius	& $r$	& $1\,\micron$\\
\end{tabular}
\caption{
Typical parameters for fluids and microscopic robots considered here. 
}\tbllabel{fluid parameters}
\tbllabel{robot parameters}
\end{table}

\section{Robot Stresses and Motion in Vessels}

We determine stresses on the robot surface by numerically evaluating the flow and robot motion in a segment of the vessel. 
For the vessel sizes, planar geometries and fluid speeds considered here, the motion and stresses can be approximated by two-dimensional quasi-static Stokes flow~\cite{hogg18}.
As examples, \fig{flow} shows the fluid velocity near the robot, in the section of the vessel indicated by the dashed rectangles in \fig{vessel geometry}. 
In spite of the different vessel geometries, the flow near the robot is similar in the two cases.

\begin{figure}
\centering 
\includegraphics[width=\widefigwidth]{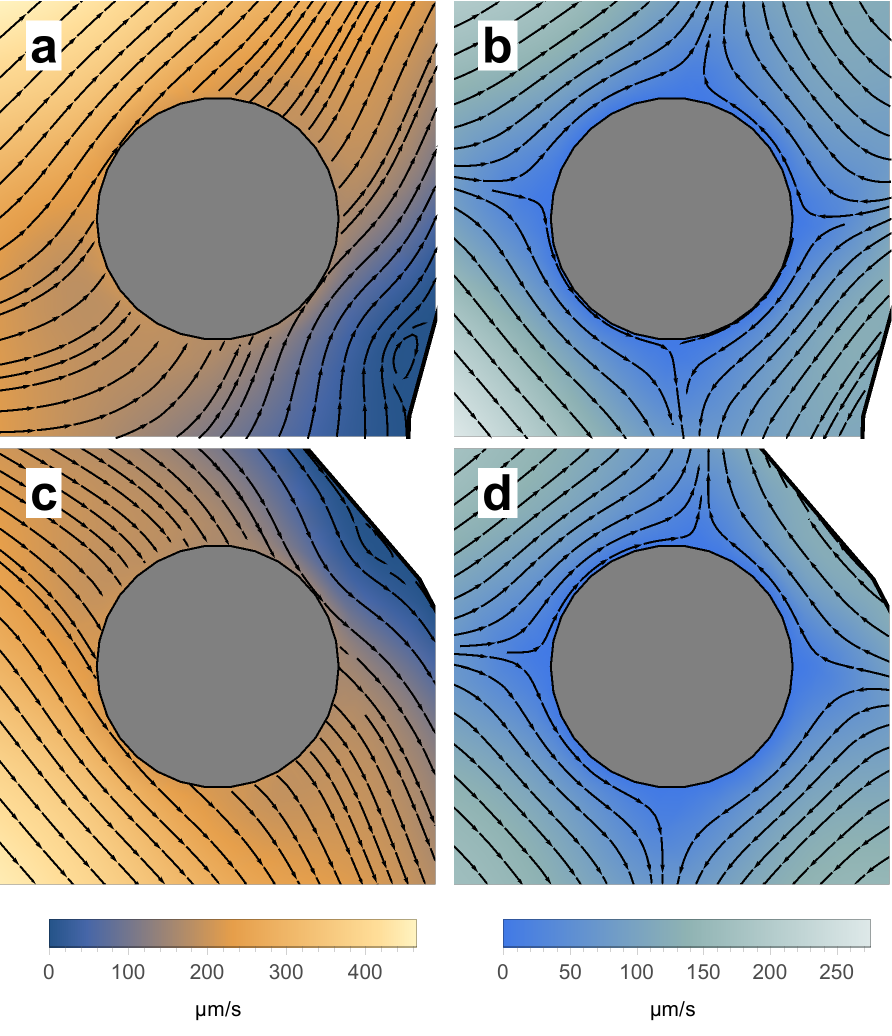}
\caption{Fluid flow near the robot for the scenarios shown in \fig{vessel geometry}.
Arrows show streamlines of the flow and colors show the flow speed. 
(a) Fluid velocity in the branch with respect to the vessel. Velocity is zero at the vessel wall, and matches the motion of the robot at its surface.
(b) Fluid velocity in the branch with respect to the robot. Velocity is zero at the robot surface.
(c) Fluid velocity in the curve with respect to the vessel.
(d) Fluid velocity in the curve with respect to the robot. 
The legend at the bottom of each column applies to the two plots above it.
}\figlabel{flow}
\end{figure}

\subsection{Stresses on Robot Surface}

\begin{figure}
\centering
\includegraphics[width=\figwidth]{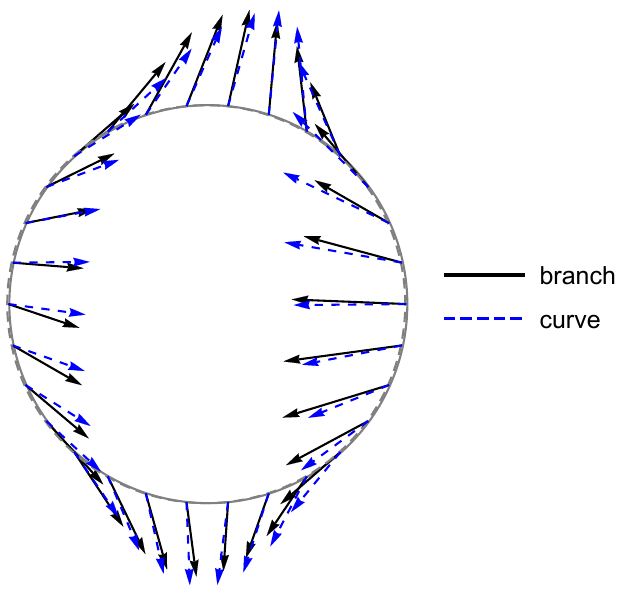}
\caption{Stress vectors on the robot surface for the indicated robot positions in the branch and curve scenarios of \fig{vessel geometry}. The arrow lengths show the magnitude of the stresses, ranging up to $0.58\,\pascal$, at locations of sensors on the robot surface.}\figlabel{stresses}
\end{figure}

We denote the stresses on a robot's surface by $\vec{s}(\theta,t)$ for the stress vector at angle $\theta$ at time $t$. The angle $\theta$ specifies a location on a robot's surface, measured from an arbitrary fixed point on the robot called its ``front''.
A robot can estimate $\vec{s}(\theta,t)$  by measuring stresses at various locations on its surface with force sensors, and interpolating between these locations. By measuring forces normal and tangential to the surface, the sensors determine the stress vector~\cite{hogg18}.

The stresses depend on the vessel geometry near the robot and the speed of the flow.
However, this relationship is not unique: different geometries can produce similar stress patterns.
\fig{vessel geometry} provides one such example. In these cases, 
\fig{stresses} shows how stresses vary over the robot's surface, with the arrows indicating the stress vectors at points spaced uniformly around the surface. The patterns of stress on the surfaces are nearly the same.
Thus, in general, the pattern of stresses at a single instant does not reliably identify the geometry of the vessel near the robot.

\subsection{Changing Stress Patterns}

As a robot moves through a vessel with changing geometry, the stresses on its surface change.
In many cases the stresses change significantly as a robot moves through a branch, whereas changes are fairly small as a robot moves around a curve. 

One measure of changing stresses is the correlation of the stress pattern at two times.
Suppose $\vec{f}(\theta)$ and $\vec{g}(\theta)$ are two vector-valued functions of the angle $\theta$ around the robot's surface. The correlation between these vector fields is
\begin{equation}\eqlabel{vector correlation}
\corr(\vec{f},\vec{g}) = \frac{1}{\norm{\vec{f}} \norm{\vec{g}}} \int_0^{2\pi} \vec{f}(\theta) \cdot \vec{g}(\theta)\,d\theta
\end{equation}
with the norm $\norm{\vec{f}} = \sqrt{ \int_0^{2\pi} \vec{f}(\theta) \cdot \vec{f}(\theta)\,d\theta}$ and $\vec{f} \cdot \vec{g}$ denoting the inner product of the two vectors.

In our case, the vector fields are the stresses on the robot surface. 
As an example, the correlation between the surface stresses in the two cases shown in \fig{stresses} is $0.98$.
Due to the normalization, the correlation is independent of the overall magnitude of the stresses. In particular, this means that the correlation does not depend on the fluid viscosity.

As noted above, the robot can estimate the stress field $\vec{s}(\theta,t)$ by interpolating surface stress measurements. A particularly useful method is interpolating the stress pattern from a few Fourier modes. The correlation between two stress patterns can be computed directly from the Fourier coefficients, avoiding the need to explicitly evaluate the integrals in \eq{vector correlation}~\cite{hogg18}. Specifically, we use the first six modes, which capture most of the variation in stress over the robot's surface for the cases considered here.

When viewed from a fixed location on the robot surface, e.g., a specific sensor, the stress changes both due to changing vessel geometry and due to the robot's rotation caused by the fluid. This rotation is not relevant for identifying vessel geometry for the spherical robots considered here. Thus, to identify changes in geometry, the robot must remove the change due to its rotation. A simple way to do so is to maximize the correlation over all possible rotations of the robot between the two measurements used in the correlation. Specifically, this approach compares the stress at time $t$, $\vec{s}(\theta,t)$, with shifted versions of the stress at a prior time, $\vec{s}(\theta+\Delta \theta, t-\Delta t)$, and uses the maximum correlation over all shifts $\Delta \theta$. That is, the robot measures changes in the pattern of stress that are \emph{not} due to its rotation by evaluating
\begin{equation}\eqlabel{max correlation}
c(t,\Delta t) = \max_{\Delta \theta} \; \corr \left( \vec{s}(\theta,t), \, \vec{s}(\theta+\Delta \theta, t-\Delta t) \right)
\end{equation}
for the correlation function of \eq{vector correlation}.
Another application of this maximization is estimating the robot's angular velocity because the shift in angle, $\Delta \theta$, giving the maximum is an estimate of how much the robot has rotated between these two times~\cite{hogg18}.

\begin{figure}
\centering
\includegraphics[width=\figwidth]{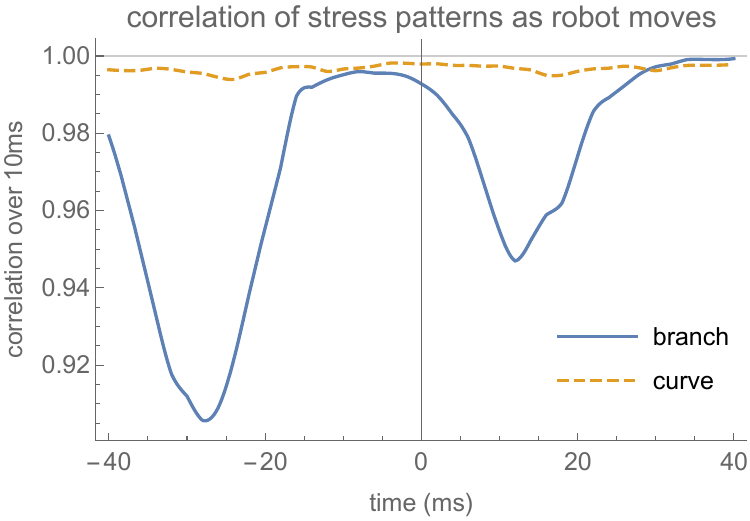}
\caption{Correlation $c(t,\Delta t)$ between stress patterns at times $t$ and $t-\Delta t$, with $\Delta t = 10\,\millisecond$, as the robot moves along the paths shown in \fig{vessel geometry}. The times correspond to the tick marks along the paths shown in \fig{vessel geometry}, with $t=0$ corresponding to the robot positions shown in that figure.
}\figlabel{correlations}
\end{figure}

As an example, \fig{correlations} shows the correlation between stress patterns separated by $10\,\millisecond$ as the robot moves through the branch and curved vessels of \fig{vessel geometry}.
For the curve, the stress pattern remains nearly the same, so correlations are close to one. For the branch, however, the correlation drops as the robot approaches the branch, about $30\,\millisecond$ before it reaches the position shown in \fig{vessel geometry}. Later, the correlation drops again as the robot moves into one of the branches.
Similarly, if the robot were moving the opposite direction, i.e., the flow was a merge of two small vessels into a larger one, the robot would encounter these drops in correlation in the opposite order and at times shifted by $10\,\millisecond$ as it compares its current stress pattern with the pattern it had encountered earlier along the reversed path.

For the flow speeds and vessel sizes considered here, $\Delta t =10\,\millisecond$ is a reasonable choice: over that time a robot can move a distance comparable to the extent of branching or curving, but not so far as to completely pass the changing geometry. 
However, the precise value of $\Delta t$ is not important. For example, comparing stresses separated by $\Delta t=5\,\millisecond$ or $20\,\millisecond$ is qualitatively similar to the behavior shown in \fig{correlations}.
For definiteness, we use $\Delta t =10\,\millisecond$ in the following discussion.

\section{Classifying Vessel Geometry}
\sectlabel{classify}

\fig{correlations} suggests a robot could identify vessel branches by checking for when the correlation of stresses separated by a short time is sufficiently small. This procedure could reliably distinguish branches from curves if there is little overlap in the distributions of minimum correlations for these two geometries. Unfortunately, the behavior shown in \fig{correlations} does not occur in all cases. Instead, the distributions of minimum correlation for curves and branches have considerable overlap, especially when the robot is near the center of the vessel.

\begin{figure}
\centering
\includegraphics[width=\figwidth]{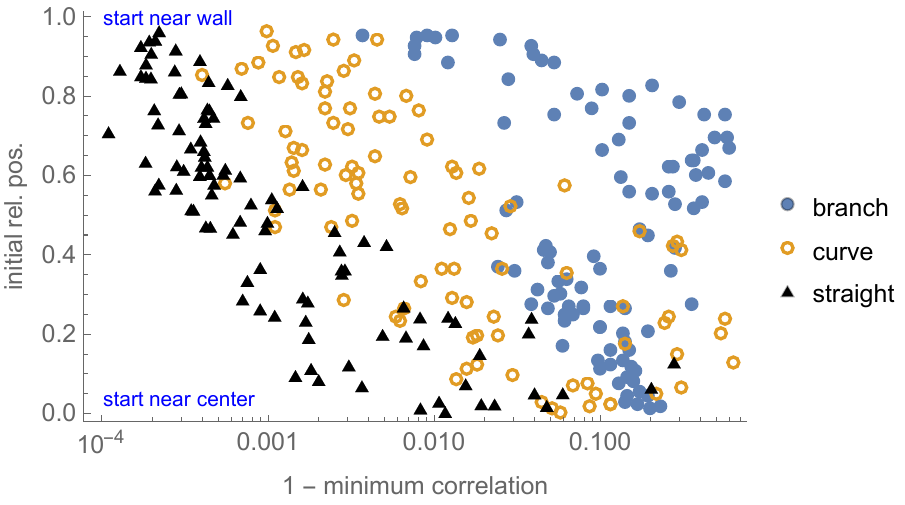}
\caption{Scatterplot of the minimum value of $c(t,\Delta t)$ along a path, for $\Delta t = 10\,\millisecond$, and initial relative position, defined by \eq{relative position}. The points distinguish robots moving through a branch, around a curve or in a straight vessel. To highlight the differences among these geometries, the horizontal axis shows $1-c$ on a logarithmic scale. Thus situations where the stress pattern changes only slightly over time $\Delta t$, i.e., correlations are close to 1, appear on the left side of the diagram.
}\figlabel{correlation and initial position}
\end{figure}

\begin{figure}
\centering
\includegraphics[width=\figwidth]{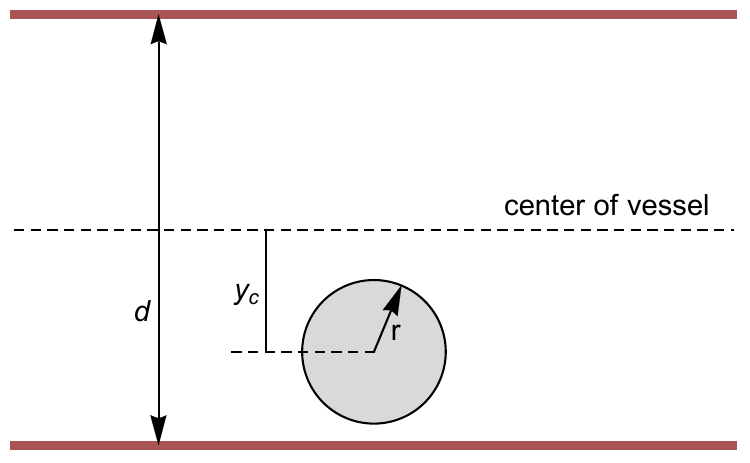}
\caption{Position of robot (gray disk) in a straight segment of a vessel, used to define the robot's relative position.}\figlabel{robot position}
\end{figure}

\fig{correlation and initial position} quantifies this difficulty by showing how the minimum correlation along a path depends on how close the robot is to the vessel wall before it reaches the curve or branch, i.e., while still in a fairly straight portion of the vessel. 
As a measure of how close a robot is to the vessel wall, we use the robot's \emph{relative position}, given by
\begin{equation}\eqlabel{relative position}
\relPos = \frac{|y_c|}{d/2-r}
\end{equation}
where $y_c$ is the position of the robot's center relative to the vessel's central axis, $d$ is the vessel diameter and $r$ the robot's radius, as illustrated in \fig{robot position}. Relative position ranges from 0, for a robot at the center of the vessel, to 1, for a robot just touching the vessel wall. 
A robot can accurately estimate its relative position while in a straight vessel segment from surface stresses~\cite{hogg18}.

When the robot starts near the center of the vessel, \fig{correlation and initial position} shows that curved paths have a wider range of minimum correlations than branches, and this range includes the values occurring in branches. Combined with smaller, and hence noisier, stresses for robots relatively far from the vessel wall~\cite{hogg18}, this indicates correlation is not a reliable identifier of branches when robots start near the vessel center.

To improve identification for paths starting close to vessel center, a third piece of information is helpful, namely a summary of the stress pattern at the time the robot evaluates the correlation. That is, to identify branches, at time $t$, we use three pieces of information: the correlation $c(t,\Delta t)$, the relative position for the path that was determined prior to any significant change in the stress, and the current stress $\vec{s}(\theta,t)$. 

To create a vessel geometry classifier using this information, we generate a set of training samples. Specifically, for each training sample (created as described in \sectA{samples}), we determine the time $t$ along the path with the minimum correlation, and use the three pieces of information from that time along the path. This method is an example of off-line training. That is, we suppose the training samples have measurements from a completed path, i.e., a path starting before the robot reaches the branch or curve, and continuing until the robot is well past those changes. With measurements along the whole path, the time of minimum correlation can be obtained and values at that time used for training.
Using such training samples from both branch and curve vessels results in the logistic regression classifier for branches described in \sectA{classify details}.

\section{Applying the Classifier to Identify Branches}
\sectlabel{classifier usage}

The classifier described above was trained with the minimum correlation along a path. A robot using the same method when applying the classifier would have to wait until it was sufficiently far past a changing geometry to be sure it had detected the minimum correlation along the path for that change. This off-line application of the classifier could be useful in reporting vessel geometry changes well after encountering them, e.g., to provide a description of the path leading the robot to a target location.

We focus on the more demanding classification task of recognizing a branch near the time the robot encounters it. This on-line or real-time classification allows the robot to take action while still near the branch. In this case, a robot uses the classifier by repeatedly evaluating \eq{p(branch) regression} as it moves. During these evaluations, the correlation $c(t,\Delta t)$ is not necessarily the minimum correlation along the path: i.e., the robot could encounter smaller values as it continues through the vessel.
Thus, the robot using this method is extrapolating beyond the values used for training.

The classifier uses the robot's relative position in the vessel before it encounters a branch or significant curve. Thus the robot must save its estimate of relative position, updating the value only while vessel geometry is not changing. A robot could determine when this steady behavior occurs by checking when the correlations between stresses at various delay times $\Delta t$ are close to 1, and the pattern of stress on its surface is consistent with its presence in a straight vessel segment~\cite{hogg18}. 
When the robot encounters changing geometry, it uses the saved estimate of its relative position when evaluating the classifier, i.e., \eq{p(branch) regression}. This procedure applies to typical capillaries~\cite{pawlik81} where significant geometry changes are separated by tens of microns, a considerably larger distance than the size of the robots small enough to pass through those vessels.

\subsection{Example}

\begin{figure}
\centering
\includegraphics[width=\figwidth]{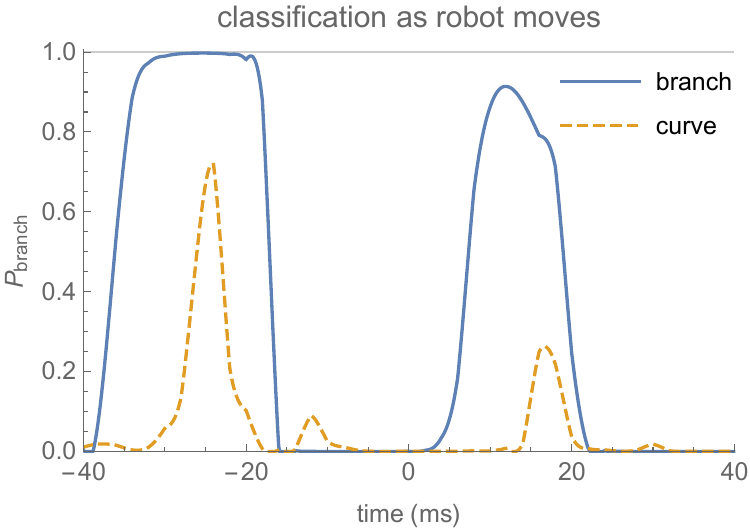}
\caption{Estimated probability of encountering a branch, \Pbranch, vs.~time as the robot moves along the paths shown in \fig{vessel geometry}, based on correlation $c(t,\Delta t)$ with $\Delta t=10\,\millisecond$.}\figlabel{classifications}
\end{figure}

As an example of how the classifier applies to on-line classification, \fig{classifications} shows the estimates of \Pbranch\ from \eq{p(branch) regression} along the robot paths of \fig{vessel geometry}. The branch path has higher values than the curve.
This example indicates how a robot could use the classifier for on-line branch identification: consider a branch to be nearby whenever \Pbranch\  exceeds a predetermined threshold. For this example, a threshold around $0.8$ distinguishes the branch from the curve.

In addition to identifying \emph{whether} the robot passes a branch, \fig{classifications} indicates \emph{when} this method detects the branch. In this case for the branching vessel, \Pbranch\ becomes large about $30\,\millisecond$ before the robot reaches the location shown in \fig{vessel geometry}. This corresponds to the robot entering the branch. The robot slows as it moves through the branch, leading to a period of large correlation for about $20\,\millisecond$. As the robot leaves the branch, the stress pattern changes again, leading to a second minimum in the correlation (see \fig{correlations}) and another maximum in \Pbranch.
In other cases, the robot moves more rapidly through the branch, so the $\Delta t=10\,\millisecond$ time difference used here gives a single minimum in the correlation and, correspondingly, a single peak in \Pbranch.
A robot could distinguish these cases by checking for a second peak within a few tens of milliseconds. Over that amount of time, the second peak indicates the robot is leaving the branch rather than encountering a second branch. This is consistent with the typical spacing of branches in capillaries (see \sect{geometry}).

\subsection{Selecting a Threshold to Identify Branches}

A suitable threshold to use for identifying branches with this classifier depends on the relative importance of false negatives (i.e., missing a branch) and false positives (i.e., incorrectly considering a curved vessel to have a branch).
The importance of these errors depends on the application. 

For example,  if a robot with locomotion capability needs to move into a branch, it is better to recognize a branch before passing it, so the robot would only need to actively move a short distance to reach the desired branch. This contrasts with the situation of not recognizing the branch until well after the robot has passed it, in which case it would need to more a larger distance and upstream against the flow of the fluid. In this case, the robot could use a relatively low threshold, thereby being fairly sure it will identify branches as it encounters them, though it may also, incorrectly, attempt to move into a branch when passing through some curved vessels. Such errors are more likely the earlier the robot needs to identify a branch because the flow well upstream of a branch is similar to that in vessel without branch.

Another action the robot could take upon detecting a branch is to move to the vessel wall near the branch and act as a beacon to other robots arriving at the branch. The beacon signal could, for example, direct subsequent robots into one branch or the other to ensure roughly equal numbers explore each branch in spite of the fluid flow favoring one branch over the other. More generally, branches could be useful locations to station robots for forming a navigation network~\cite{freitas99}. With a limited number of robots to form this network and if it is sufficient to station robots at some rather than all branches, the robot could use a high threshold to ensure identified branches are very likely to actually be branches.

Other applications have less need for identifying branches while robots are close to them, but instead collect information on the number and spacing of branches for later use, e.g., as a map or diagnostic tool at larger scales than short segments of a single vessel. This applies to identifying vessel geometries that distinguish different organs or normal from cancer tissue~\cite{nagy09,jain14}.
In such cases, the emphasis could be on identifying all branches, favoring a relatively low threshold, and then verifying detected branches with subsequent measurements after the robot has moved past each possible branch, thereby reducing the false positives while keeping sensor information obtained near the branches from the original identification of those cases later verified to be branches.

Multiple robots spaced closely enough could communicate verified branch detections to robots upstream of the branch. 
With a message from a downstream robot that it verified its recent passage of a branch, a robot could lower its threshold in anticipation of that upcoming branch. The message could come from a downstream robot that entered a different branch of the splitting vessel than the robot receiving the message.
For example, with acoustic communication robots could compare measurements over $100\,\micron$ or so~\cite{hogg12}, which is farther than the typical spacing between branches in capillaries described in \sect{geometry}.

\subsection{Verification After Passing a Curve or Branch}

This paper describes how a robot can use stresses to identify branches as it encounters them. This contrasts with off-line methods that collect time-stamped information before, during and after a robot passes a branch, and later use the entire path history to identify if and when the robot passed branches. A possible combination of the two approaches is to use on-line classification to identify likely branches, record information about them, and later verify the branch detection when the robot has additional information after passing the possible branch.

Information a robot could use for this verification task includes changes in vessel diameter, relative position and robot speed before and after passing the curve or branch. Typically, these changes are much larger after passing a branch than a curve.
Such comparisons must wait until the robot is well past the branch or curve, when it is again in a nearly straight vessel segment where stress-based navigation allows estimating these quantities~\cite{hogg18}.

Specifically, when a vessel splits into branches, the branches have smaller diameter than the main vessel, but the combined cross sections of the branches is larger than that of the main vessel~\cite{murray26}. Thus measuring vessel diameter before and after the branch gives a direct indication of branching.

The increased cross sectional area after a vessel splits leads to slower flows in the branches~\cite{sochi15}, whereas flow in a curve remains nearly constant. This suggests a robot could check for changes in its speed through the vessel before and after passing a branch or curve to verify the identification.
However, slower fluid speed in the branch does not necessarily mean the robot's speed changes in the same proportion, because the robot's speed depends both on the speed of the flow and how close the robot is to the wall, i.e., its relative position. Along with flow speed, the relative position can change as a robot passes a branch.

\fig{3paths} illustrates this behavior. Robots close the wall remain close upon entering a branch or moving around a curve. But robots entering a branch near the center of the vessel move to near the wall of the branch, and robots starting between the center and the wall move to near the middle of a branch.
This change in relative position can distinguish branching from curved paths that are not close to the wall. 
Similarly, the change in speed is particularly large for branch paths when the robot is not near the wall, so its speed reflects the decreasing flow through the branches. When the robot is near the wall, in either a branch or curve, it moves relatively slowly and remains near the wall, giving relatively little change.

\begin{figure}
\centering
\includegraphics[width=\widefigwidth]{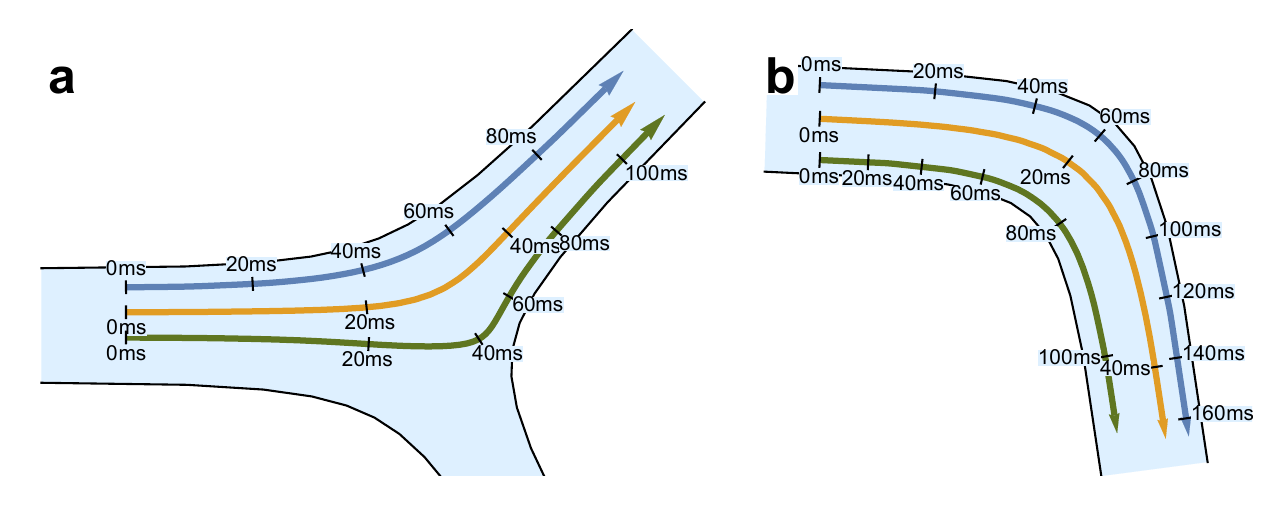}
\caption{Robot paths through (a) branched and (b) curved vessels. Each path is for a robot traveling by itself through the vessel with the fluid. The ticks along the paths indicate times, in milliseconds, from the start of each path when the maximum inlet flow speed is $1000\,\micron/\second$. Robots along paths near the center of the vessel move more rapidly than those near the walls, indicated by the wider spacing between successive ticks along the central paths. The vessel inlets both have diameters of $7.8\,\micron$.}\figlabel{3paths}
\end{figure}

Changes in relative position and speed are particularly useful for distinguishing curves and branches for paths near the middle of the vessels, precisely the cases where the correlation-based method in this paper are least reliable (see \fig{correlation and initial position}). Moreover, the larger differences in these measures between curves and branches for paths near the center of the vessel could compensate to some extent for the lower accuracy of estimating position and speed from stresses when the robot is near the center of the vessel~\cite{hogg18}.

Due to errors in estimating position, speed and vessel diameter from stresses~\cite{hogg18}, evaluating changes from a combination of these measures before and after passing a branch or curve is likely to be more robust than relying on a single method.

\section{Classification Performance}
\sectlabel{accuracy}

This section evaluates how the classifier performs using a set of test samples.

In the branch vessels considered here, the fluid flows from the larger vessel into the two branches. Thus, testing the classifier with these paths evaluates how well a robot recognizes when the vessel splits into two smaller vessels, with the robot moving into one of them.
The situation for flow merging from smaller branches into a larger vessel is similar. This is because for a robot at a given location in a vessel, stresses on the robot surface are the same for either direction of the flow due to the reversibility of Stokes flow~\cite{happel83}. Thus the stress measurements along a path are the same for paths through merging branches, but in the reverse order. 

For merging vessels, the classification operates in the same way as for splitting vessels, but with a shift of time $\Delta t$ in when the correlation is measured. 
For instance, for the robot at the location shown in \fig{vessel geometry}a, the correlation $c(t,\Delta t)$ compares the stresses at the robot's indicated location with the stresses when the robot's center was at the point along the path indicated by the tick for $-10\,\millisecond$ when using $\Delta t=10\,\millisecond$.
Conversely, a robot moving in the reverse direction along the path, i.e., from the branch at the upper right into the main vessel at the left, would compare its stress with that $10\,\millisecond$ earlier on the reverse path, corresponding to the location of the tick for $10\,\millisecond$ on the forward path. Thus while stresses are the same along the forward and reversed paths, the comparison used for the correlation and the initial position in the vessel before encountering the branch are different for these two directions.

When the robot along the reverse path is at the position indicated by the tick for $-10\,\millisecond$, it would compare stresses at the same two times that the robot on the forward path does at the position indicated in the figure. So a robot on the reverse path computes the same correlations as a robot on the forward path, but with a shift of $\Delta t$ in time.
In this comparison, the robots use the same correlation in the classifier. But because they are at different locations along the path when they evaluate that correlation, they would have different values for the other two values used in the classifier: the current stress and the initial relative position.

The classifier training only included the forward direction for each branch, i.e., moving from a main vessel into one of the branches at a split. As a test of how well the classifier generalizes, we use both path directions for each branch test sample.

\subsection{Accuracy}

Since the choice of threshold depends on the application, instead of characterizing a classifier for a single choice of threshold, a better measure is the tradeoff between true and false positives over the range of threshold values. At one extreme, a threshold equal to one means the robot never recognizes a branch (no true positives) but also never mistakenly considers a curve to be a branch (no false positives). At the other extreme, a threshold equal to zero means the robot always considers itself to be encountering branches: this identifies all actual branches, but also mistakenly recognizes curves and straight vessel segments as branches (high false positives). For a perfect classifier, there would exist an intermediate threshold allowing it to identify all branches without also mistaking other geometries for branches.

For the classifier developed in this paper, \fig{tradeoff} shows the performance as the threshold varies from 1 (at lower left) to 0 (at upper right).
Specifically, for each threshold value, the figure evaluates \Pbranch\ with \eq{p(branch) regression} along the path of each test sample. If \Pbranch\ exceeds the threshold anywhere along the path, that sample is classified as a branched vessel.
The true positives are the branch samples identified as branches using that threshold, and the false positives are the curve samples incorrectly identified as branches.

This classifier performs well: with suitable threshold, the classifier recognizes most branches with only a few mistakes, as indicated by the curve in \fig{tradeoff} passing close to the ideal behavior of recognizing all true positives and none of the false positives (i.e., the upper left corner of the figure).
This tradeoff curve includes both forward and reverse paths. Examining performance separately for each direction (i.e., splitting or merging vessels) shows similar curves. Thus there is no penalty for not including reverse paths when training the classifier. 
This is an indication of the robustness of the information used for this classifier, and the simplicity from the reversibility of Stokes flow.

An overall performance measure from the tradeoff of \fig{tradeoff} is the area under the curve. In this case, the area is $98.6\%$ of the total area. 
By comparison, the area would be $50\%$ for a classifier that made no distinction between branch and curve, and $100\%$ for a perfect classifier.

\begin{figure}
\centering
\includegraphics[width=\figwidth]{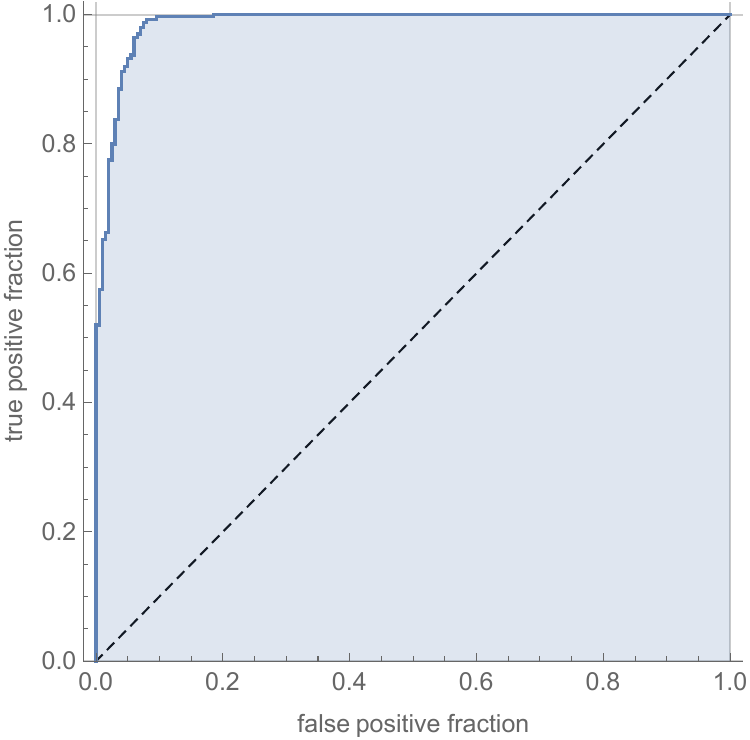}
\caption{Classification performance on test samples. The curve would follow the dashed diagonal line if the classifier did not discriminate between paths through branch and curve vessels.}\figlabel{tradeoff}
\end{figure}

\subsection{When Branches Are Identified}

\begin{figure}
\centering
\includegraphics[width=\figwidth]{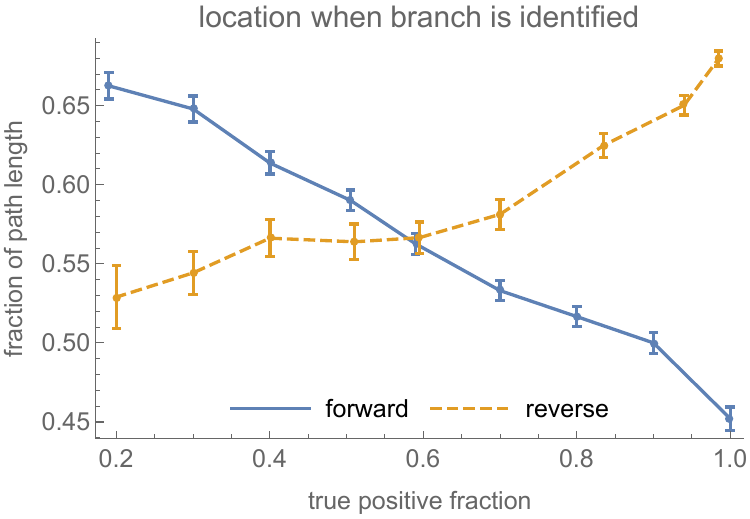}
\caption{Fraction of path length of the test sample paths at which \Pbranch\ first exceeds the detection threshold corresponding to the true positive fraction indicated on the horizontal axis. The solid and dashed curves are for forward and reverse branch paths, respectively. Error bars show the standard error of the means, indicated by the points.}\figlabel{branch detection}
\end{figure}

In addition to how accurately this classifier detects branches, an important performance measure is \emph{where} along a path the robot first detects a branch. One of the features this classifier uses is the correlation between stresses at the robot's current location and those at the time $\Delta t = 10\,\millisecond$ earlier. Typically, stresses change the most as the robot passes through the branch, and during $10\,\millisecond$ the robot does not move significantly past the branch. This means the classifier typically detects the branch while the robot is within the branching section of the vessel.

Quantitatively, \fig{branch detection} shows this classifier generally identifies branches near the middle of the sample paths. This corresponds to when the robot is close to the branch, rather than well before or after passing the branch. That is, this classifier identifies branches while the robot is close to them. Thus a robot can use this stress-based classifier to identify when it is passing a branch, well before it passes significantly downstream of the branch. On the other hand, branches are not identified well before the robot reaches the branch. Thus this classifier is not useful for applications requiring significant advance notice that the robot is approaching a branch. 

This classifier can detect most branches (high true positive fraction) with only a few false positives (see \fig{tradeoff}). Thus applications will likely use thresholds low enough to give true positive fraction above, say, 80\% or so, corresponding to the right portion of \fig{branch detection}. With this choice for the threshold, branch detection will generally occur at smaller fractions of the path length for forward paths than for reverse paths. This corresponds to a robot moving toward a vessel split detecting the branch just as the main vessel splits. Conversely, a robot moving in a vessel that merges with another to form a larger vessel will identify the branch just as the vessels merge.
Quantitatively, the difference path fraction for these two cases shown in the figure (about $0.15$ for thresholds giving true positives above $80\%$), corresponds to a difference of about $6\,\micron$ for the path lengths used here (see \sectA{samples}). Thus the difference in where a branch is first identified for splitting or merging vessels is just a few times the robot diameter.

\subsection{Noise}

\begin{figure}
\centering
\includegraphics[width=\figwidth]{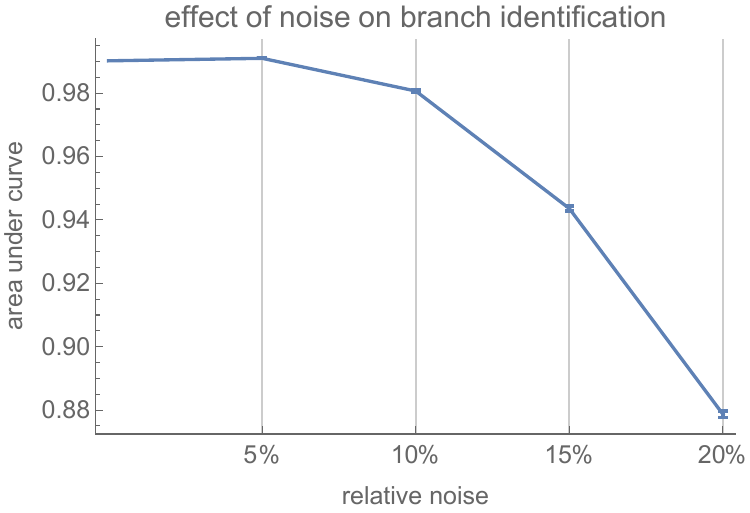}
\caption{Area under the tradeoff curve as a function of relative noise added to classifier inputs. Each value is the average of 100 noisy versions of the measurements along paths of all the test samples. The error bars are the standard error of those averages, and are only slightly larger than the thickness of the curve connecting the points.}\figlabel{noise}
\end{figure}

Sensor measurements are subject to noise, so in practice robots will not have exact values for the stress-based quantities used by the classifier. Evaluating the effect of noisy measurements on classification accuracy is an important performance measure.

The classifier uses three pieces of information: correlation of two stress measurements separated by $\Delta t$, relative position estimated from stress while in a straight segment, and principal components from the latest stress used for correlation. These quantities involve stress measurements at different times, except that one of the stresses used for correlation is the same as used to determine principal components. Thus a reasonable model of the noise is independent errors for the three values used as inputs by the classifier. This assumes noise at different times, separated by at least a few milliseconds, is uncorrelated, e.g., as is the case for thermal noise.

As one example of the effect of independent noise, \fig{noise} shows the consequence of relative errors for the area under the tradeoff curve for true and false positives shown in \fig{tradeoff}. For simplicity, the noise used in this figure takes each of the three inputs to the classifier to have a noisy value given by a random relative error drawn from a normal distribution with zero mean and standard deviation equal to the amount of relative noise indicated on the horizontal axis of the figure. This shows little change in classifier performance with up to about $10\%$ noise. More generally, each of the three inputs to the classifier could have a different amount of noise, depending on how noise from stress measurements propagates to the calculated values of correlation, relative position and principal components. 

If noise is large enough to significantly degrade classifier performance, the robot could average the result of several subsequent evaluations. For instance, averaging stress measurements over several milliseconds can significantly reduce the effect of thermal noise on stress measurements~\cite{hogg18}. This averaging will also improve the stress-based inputs to the classifier, provided noise at different times is independent.

Averaging to reduce the effect of independent noise contrasts with systematic errors, which produce noise correlated over time or across different sensors. Examples of such systematic errors include a stress sensor giving consistently smaller readings than the others due to poor calibration, or if sensor responses change as the robot moves through the vessel due to sensors becoming covered by material adhering to the robot surface. Another source of systematic error is a robot's clock running at the wrong speed. That could, for instance. cause a robot to actually compare stresses separated by $\Delta t=15\,\millisecond$ while using a classifier trained with correlations with $\Delta t=10\,\millisecond$. Such systematic errors are not reduced by averaging so will need to be handled by designing stable sensors or providing a way to calibrate them while in use. Alternatively, if such errors differ among robots, e.g., because they have different clock speeds, nearby robots could communicate the information they use for branch identification. Each robot could then average its own measurement with those of other robots to reduce the effect of these errors.

\section{Discussion}

This paper describes how a robot moving with fluid in a small vessel can use the changing stress on its surface to identify when it encounters a branch. Repeatedly using this classifier, a robot can identify changes in the vessel geometry as it moves. For example, the robot could record the time and hence distance, with an estimate of its speed obtained, e.g., from stress measurements~\cite{hogg18} between between its initial approach and final passing of a branch. Thus the robot can determine the size of the branching region. Over longer times, the robot could measure distances between successive branches, thereby estimating the larger-scale geometry of successive splits and merges of small vessels.

The classifier in this paper uses only a few properties of the changing stress pattern. These properties are sufficient to identify branches for the scenarios considered here.
It remains to be seen how well these properties perform in more complex situations. 
One such situation is nonplanar branching vessels~\cite{sochi15} which require evaluating 3D flows.
Another situation is the fluid containing a variety of objects in addition to the robot. Such objects include cells that deform as they are pushed around curves or into branches, requiring more elaborate simulations~\cite{hoskins09}, especially in somewhat larger vessels than considered here~\cite{bagchi07}.
In these or other more complex situations, the properties used in the classifier described here may not be sufficient for high accuracy. In such cases, classifiers could incorporate additional information available to the robot about its local environment.
These additions include prior information about vessel structure and additional measurements.

For instance, the robot could combine the classifier's output with prior probabilities of branches. This prior could, for instance, consider typical distances between successive branches and how often vessels bend without branching. 
Values for the priors could come from studies of microvasculature in general or in specific organs~\cite{augustin17} or types of tissue~\cite{nagy09,jain14}.
Using priors specific to individual organs or tissues would require the robot to determine which organ or type of tissue it is in, e.g., using chemical signatures or external localizing signals~\cite{freitas99}.
A robot could also adjust its priors based on its recent history, e.g, using the frequency of verified branches it has encountered recently in small vessels. 
The robot can combine any prior probabilities it has with the result \Pbranch\ from the classifier using Bayes theorem.
For example, if branches are rare compared to curves, the robot could increase the threshold it uses to identify a branch to reduce the number of false positives. This would adjust for the use of an equal number of curve and branch paths used for the training in this paper, which corresponds to an equal likelihood for the robot to encounter a curve or branch.

The robot could use a wider variety of values to aid classification in more complex scenarios.
For instance, the classifier of this paper evaluates stress changes over a fixed interval of time, namely $10\,\millisecond$.
More generally, robots could use stress correlations over multiple time intervals, e.g., using $\Delta t$ values of both $10\,\millisecond$ and $20\,\millisecond$. A mix of times could help identify vessel structure when flow speed varies over a larger range than considered here, e.g., due to vessel blockages.

Another example of additional information the robot could use is a broader measure of how stress changes. The correlation measure used here measures the change in the \emph{shape} of the stress pattern on the robot surface.
As a robot moves, the stress magnitudes change as well. For instance, moving into a branch with slower flow typically decreases the stress, whereas moving closer to the wall increases it. Thus another measure of changing stress is ratio of current and previous average stress magnitudes over the surface, which characterizes the change in the \emph{magnitude} of stresses.
For the cases considered here, the change is stress magnitude gives similar information as that obtained from the correlation, and does not noticeably improve classifier performance. Nevertheless, in other cases, changes in shape and magnitude may provide different information, thereby improving performance when used together.

In addition to using the information a robot obtains while passively moving with the fluid, robots could actively probe their environments. One such approach is making changes to the surface that affect stresses.
For instance, a robot that can alter its shape~\cite{castano00} could measure how its shape affects its surface stresses to gain additional information on the nearby vessel geometry. 
Another possibility is for a robot with locomotion capability to alter its distance to the vessel wall before it encounters a branch or curve to increase the accuracy of classification or verification after passing the change. A robot could move closer to the center of the vessel so its distance to the wall will change much more when encountering a branch than a curve (see \fig{3paths}) thereby improving verification. Alternatively, it could move closer to the wall so the change in correlation is more distinct between branches and curves (see \fig{correlation and initial position}).

Beyond classifying the type of geometry, e.g, a branch, a robot might use stresses to estimate quantitative properties of how the geometry changes. This possibility arises from stresses depending on geometric properties such as the branching angle when a vessel splits and the size of the branches, as well as the radius of curvature of a curved vessel. 
Estimating such properties requires determining how they influence the stress pattern as the robot moves, and then determining how to extract property values from these influences on measured stresses. Such estimates could also benefit from prior information on relations between branching angles and diameters~\cite{thompson92}.

Communication among nearby robots could enhance these extensions. In particular, when multiple communicating robots pass through a vessel at around the same time, they could compare stresses at different positions in the vessel. These comparisons could include robots close to the wall or close to the middle, and robots approaching, near and past a branch. More broadly, robots with a variety of sensors could combine information from stresses with other measures, such as changing chemical concentrations or acoustic echos~\cite{freitas99}, to evaluate the changing geometry of their environments.


In summary, a robot can use fluid stresses on its surface to identify when it encounters branches in microscopic vessels. 
The classifier discussed here combines three types of information derived from stresses: how stress patterns change over a short time, the position of the robot in the vessel before it encounters a branch or curve, and the shape of the stress pattern. Together, these values accurately distinguish branches from curves for a typical range of geometries encountered with microscopic vessels in tissue. Moreover, the time over which stresses are compared, e.g, $10\,\millisecond$, is short enough that the robot moves just a few times its size during the evaluation. This allows the classifier to identify branches while the robot is still near the branch.

Fabrication of microscopic robots with stress sensors and even relatively simple computers involve significant technological challenges. 
Prior to the feasibility of robot fabrication, theoretical studies, such as presented in this paper, can quantify likely robot capabilities and their performance in various applications. These studies can be useful guides to the kinds of applications robots will be able to perform as their capabilities improve. Conversely, these studies quantify the performance requirements of the robot sensors and computers needed to address these applications.

\section*{Acknowledgements}

I thank Robert A. Freitas~Jr., Ralph C. Merkle and James Ryley for helpful discussions.

\newpage
\appendix

\numberwithin{equation}{section}
\numberwithin{figure}{section}
\numberwithin{table}{section}

\begin{center}
{\huge \textbf{Appendices}}
\end{center}

\section{Samples of Robot Motion in Small Vessels} 
\sectlabel{samples}

We determine the relationship between surface stresses and vessel geometry from a set of samples in known geometries.
If robots could be fabricated, these samples could be obtained experimentally, e.g., by measuring forces on microfluidic devices~\cite{wu10}. 
Since this experimental approach is not yet feasible, we instead create samples from numerical solution of the flow with a robot in vessels with various geometries.

We create samples of robot paths for a range of vessel diameters and flow speeds corresponding to small blood vessels, with parameters given in \tbl{fluid parameters}. 
These samples are variations of the situation shown in \fig{vessel geometry} with parameters chosen uniformly at random according to:
\begin{itemize}

\item Maximum inlet fluid speed between 800 and $1000\,\micron/\second$. 
 
\item For curved vessels:
\begin{itemize}
\item Vessel diameter between 6 and $13\,\micron$.
\item Bend angle, between direction of inlet and outlet, between $25^\circ$ and $75^\circ$.
\end{itemize}

\item For a vessel splitting into two branches:
\begin{itemize}
\item Diameters of the branches, $d_1$ and $d_2$, between 6 and $10\,\micron$.
\item The diameter of the main vessel, $d$, is determined from $d_1$ and $d_2$ according to Murray's law~\cite{murray26,sherman81,painter06}, i.e., $d^3=d_1^3+d_2^3$.
\item The two branch angles chosen uniformly between $25^\circ$ and $75^\circ$ and $-25^\circ$ and $-75^\circ$, respectively.
\end{itemize}

\item Vessel segment length extending $30\,\micron$ in each direction from the curve or branch.

\item Initial position of robot's center: eight times the robot radius, i.e., $8r=8\,\micron$, from the vessel inlet, and randomly positioned between the vessel's walls with minimum gap $0.2r$ between the robot surface and the wall.

\item Robot orientation between 0 to 360 degrees.
\end{itemize}
From the initial robot position, we solve for the robot's motion through the vessel until it comes within $8\,\micron$ of an outlet.
Boundary conditions on the flow are a parabolic velocity profile at the inlet, no-slip along the vessel wall and zero pressure at the outlet for a curve, or at both outlets for a branch.

This study used a total of 2000 samples created according to this procedure, with 1000 for each vessel type, i.e., curve or branch.
For each of these vessel types, 800 samples were used for training the classifier and the remaining 200 were used to test the classifier performance.
The paths in these samples are about $40\,\micron$ in length. The fluid typically moves the robot along the path in around $100\,\millisecond$.

\section{Identifying Branches From Stress Measurements}
\sectlabel{classify details}

This section describes how the vessel branch classifier was trained and the parameters of the resulting model. The section also estimates the computational requirements for using the trained classifier.

\subsection{Regression Classifier for Branch Detection}

This paper identifies branches with a logistic regression based on three characteristics of the robot's stresses along its path through a vessel. 
First, the correlation between the current stress and that of a short time earlier. 
Second, the robot's relative position in the vessel 
evaluated during the robot's most recent passage through a nearly straight section of the vessel. The third characteristic is a measure of the shape of the robot's current stresses, specifically the first principal component of the Fourier coefficients of the stress pattern~\cite{hogg18}.

The regression model for the probability of a branch, \Pbranch, is
\begin{equation}\eqlabel{p(branch) regression}
\Pbranch= \frac{1}{1+\exp \left(-b(\log(1-c), \relPos, p_1) \right)}
\end{equation}
where $c$ is the correlation between changing stress patterns, defined in \eq{max correlation}, \relPos\ is the relative position, defined in \eq{relative position}, $p_1$ is the first principal component of the stress pattern, and
\begin{equation*}
b(\lc,\relPos,p_1) = \beta_0 + \beta_{1} \, \lc +  \beta_{2} \, \relPos + \beta_{1,1} \, \lc^2 + \beta_{2,2} \, \relPos^2 + \beta_{3} \, p_1
\end{equation*}
where the $\beta_{\ldots}$ are the parameters, given in \tbl{p(branch)}, determined from the training samples.

Training this regression used stress measurements along the complete path of each sample to identify the time with minimum correlation along the path. This ``off-line'' training corresponds to the situation after robots complete their paths, so stress measurements all along the paths are available for training. Specifically, for each training sample, $\relPos$ is the relative position at the start of the sample path where, by construction, the robot is in a straight vessel segment prior to reaching the curve or branch. Moreover, $c$ is the minimum correlation along the path, i.e., the minimum value of $c(t,\Delta t)$ along a path, for $\Delta t = 10\,\millisecond$. The branch and curve points in \fig{correlation and initial position} are examples of these relative position and correlation values. Finally, the value of $p_1$ used for training is the principal component of the robot's stress at the time of the minimum correlation.

\begin{table}
\centering
\begin{tabular}{ccc}
parameter			& value	& standard error \\ \hline
$\beta_0$			& $-0.8$	& $0.7$ \\
$\beta_1$			& $-1.7$	&  $0.6$ \\
$\beta_2$			& $9.0$	&  $1.8$ \\
$\beta_{1,1}$		& $-0.97$	&  $0.11$ \\
$\beta_{2,2}$		& $6.8$	&  $2.3$ \\
$\beta_3$			& $11.6$	&  $0.8$ \\
\end{tabular}
\caption{Regression parameters for the probability a robot encounters a branch, \eq{p(branch) regression}.}\tbllabel{p(branch)}
\end{table}

\subsection{Computational Requirements}
\sectlabel{computation}

An important metric for classifiers is the computational cost to train and use them. The training considered here is off-line, from stress measurements collected along a sample of paths. In this case, training could take place in a conventional computer rather than in the robots, and the resulting regression parameters stored in the robot's memory. Thus training is not significantly constrained by the robots' on-board computational capabilities. 

On the other hand, robots applying the classifier would use their on-board computer to repeatedly evaluate the trained classifier from their stress measurements. Microscopic robots are likely to have limited on-board computational capabilities. Thus, it is important to evaluate the computational cost for robots using the classifier, and the memory required to store the parameters obtained from the training.

The classifier used in this paper involves a small set of parameters (see \tbl{p(branch)}), and recording several sets of stress measurements (represented by their low-frequency Fourier coefficients) to allow evaluating correlations. This information amounts to about 100 numbers, which could fit in a kilobyte of memory. The regression parameters are just a few additional numbers, so do not add significantly to the memory requirement.

The classifier uses simple functions of stress measurements. 
Specifically, \eq{p(branch) regression} involves correlations, estimates of a robot's relative position and the principal component of Fourier modes of the stresses. All these quantities can be computed from a few low-frequency Fourier coefficients of stress measurements on the robot's surface. A computer capable of $10^6\,\text{operations}/\second$ could evaluate these values every few milliseconds~\cite{hogg18}.

Such computation is well beyond the demonstrated capabilities of current nano-scale computers, e.g., DNA or RNA-based logic operations~\cite{douglas12,green17} or programmable microorganisms~\cite{ferber04}. However, theoretical analyses indicate more elaborate molecular computers could be both small enough to fit within micron-size robots and readily exceed this estimate of the required computational performance~\cite{drexler92,merkle18}. Though such computers cannot yet be manufactured, the same technology required to build the micron-scale robots considered in this paper is also likely to have the capability of fabricated these computers. Thus whenever these robots can be manufactured, they likely will have sufficient computing capability to apply the stress-based estimates described in this paper. Nevertheless, early versions of such robots may not have sufficient computation to evaluate the classifier, but still have enough to encode stress measurements for communication. In that case, an alternate method for applying the classifier would be to send stress measurements to an external computer that would evaluate the classifier and return the value to the robot. This illustrates how microscopic robots could rely on on-board computation or communication, depending on which is easier to manufacture.

The classifier considered here relies on the correlation between stresses separated by $\Delta t=10\,\millisecond$. Similar classifiers could be trained with other values of $\Delta t$. However, the change in stress and hence the correlation values depend on $\Delta t$: generally, stresses change more over longer times, leading to smaller correlations. Thus, to use a classifier trained with a specific value of $\Delta t$ requires the robot measure correlations over approximately the same time interval. For the classifier considered here, this requires the robot have a clock with, roughly, millisecond precision over time intervals of tens of milliseconds.
An alternative to on-board clocks is an external timing signal~\cite{freitas99}. For example, timing signals could use sound waves, which travel at about $1500\,\meter/\second$ in water and tissue. Thus a millisecond-period timing signal would travel $1.5\,\meter$, thereby providing a common millisecond time standard to robots operating within a few centimeters. Such an external timing signal would not only provide timing for individual robots, but also a global standard for a group of robots, allowing them to synchronize measurements.




\clearpage

\begin{thebibliography}{10}

\bibitem{allen04}
Thresa~M. Allen and Pieter~R. Cullis.
\newblock Drug delivery systems: Entering the mainstream.
\newblock {\em Science}, 303:1818--1822, 2004.

\bibitem{andrews18}
Lauren~B. Andrews, Alec A.~K. Nielsen, and Christopher~A. Voigt.
\newblock Cellular checkpoint control using programmable sequential logic.
\newblock {\em Science}, 361:1217, 2018.

\bibitem{augustin17}
Hellmut~G. Augustin and Gou~Young Koh.
\newblock Organotypic vasculature: From descriptive heterogeneity to functional
  pathophysiology.
\newblock {\em Science}, 357:eaal2379, 2017.

\bibitem{bagchi07}
Prosenjit Bagchi.
\newblock Mesoscale simulation of blood flow in small vessels.
\newblock {\em Biophysical Journal}, 92:1858--1877, 2007.

\bibitem{bleckmann09}
Horst Bleckmann and Randhy Zelick.
\newblock Lateral line system of fish.
\newblock {\em Integrative Zoology}, 4:13--25, 2009.

\bibitem{bouffanais10}
Roland Bouffanais, Gabriel~D. Weymouth, and Dick K.~P. Yue.
\newblock Hydrodynamic object recognition using pressure sensing.
\newblock {\em Proc. of the Royal Society A}, 467:19--38, 2010.

\bibitem{cassot06}
Francis Cassot et~al.
\newblock A novel three-dimensional computer assisted method for a quantitative
  study of microvascular networks of the human cerebral cortex.
\newblock {\em Microcirculation}, 13:15--32, 2006.

\bibitem{castano00}
A.~Castano, W.~M. Shen, and P.~Will.
\newblock {CONRO}: Towards miniature self-sufficient metamorphic robots.
\newblock {\em Autonomous Robots}, 8:309--324, 2000.

\bibitem{douglas12}
Shawn~M. Douglas, Ido Bachelet, and George~M. Church.
\newblock A logic-gated nanorobot for targeted transport of molecular payloads.
\newblock {\em Science}, 335:831--834, 2012.

\bibitem{drexler92}
K.~Eric Drexler.
\newblock {\em Nanosystems: Molecular Machinery, Manufacturing, and
  Computation}.
\newblock John Wiley, NY, 1992.

\bibitem{dusenbery09}
David~B. Dusenbery.
\newblock {\em Living at Micro Scale: The Unexpected Physics of Being Small}.
\newblock Harvard Univ. Press, Cambridge, MA, 2009.

\bibitem{eberhardt16}
W.~C. Eberhardt et~al.
\newblock Development of an artificial sensor for hydrodynamic detection
  inspired by a seal's whisker array.
\newblock {\em Bioinspiration \& Biomimetics}, 11:056011, 2016.

\bibitem{ferber04}
Dan Ferber.
\newblock Microbes made to order.
\newblock {\em Science}, 303:158--161, 2004.

\bibitem{freitas99}
Robert~A. {Freitas Jr.}
\newblock {\em Nanomedicine}, volume {I}: Basic Capabilities.
\newblock Landes Bioscience, Georgetown, TX, 1999.
\newblock Available at www.nanomedicine.com/NMI.htm.

\bibitem{green17}
Alexander~A. Green et~al.
\newblock Complex cellular logic computation using ribocomputing devices.
\newblock {\em Nature}, 548:117--121, 2017.

\bibitem{happel83}
John Happel and Howard Brenner.
\newblock {\em Low {Reynolds} Number Hydrodynamics}.
\newblock Kluwer, The Hague, 2nd edition, 1983.

\bibitem{hill08}
Ciaran Hill, Antonio Amodeo, Jean~V. Joseph, and Hitendra R.~H. Patel.
\newblock Nano- and microrobotics: how far is the reality?
\newblock {\em Expert Review of Anticancer Therapy}, 8:1891--1897, 2008.

\bibitem{hogg18}
Tad Hogg.
\newblock Stress-based navigation for microscopic robots in viscous fluids.
\newblock {\em J. of Micro-Bio Robotics}, 2018.

\bibitem{hogg12}
Tad Hogg and Robert~A. {Freitas Jr.}
\newblock Acoustic communication for medical nanorobots.
\newblock {\em Nano Communication Networks}, 3:83--102, 2012.

\bibitem{hoskins09}
M.~H. Hoskins, R.~F. Kunz, J.~E. Bistline, and C.~Dong.
\newblock Coupled flow-structure-biochemistry simulations of dynamic systems of
  blood cells using an adaptive surface tracking method.
\newblock {\em J. of Fluids and Structures}, 25:936--953, 2009.

\bibitem{jager00}
Edwin W.~H. Jager, Olle Inganas, and Ingemar Lundstrom.
\newblock Microrobots for micrometer-size objects in aqueous media: Potential
  tools for single-cell manipulation.
\newblock {\em Science}, 288:2335--2338, 2000.

\bibitem{jain14}
R.~K. Jain, J.~D. Martin, and T.~Stylianopoulos.
\newblock The role of mechanical forces in tumor growth and therapy.
\newblock {\em Annual Review of Biomedical Engineering}, 16:321--346, 2014.

\bibitem{ke18}
Yonggang Ke, Carlos Castro, and Jong~Hyun Choi.
\newblock Structural {DNA} nanotechnology: Artificial nanostructures for
  biomedical research.
\newblock {\em Annual Reviews of Biomedical Engineering}, 30:377--403, 2018.

\bibitem{koman18}
Volodymyr~B. Koman et~al.
\newblock Colloidal nanoelectronic state machines based on {2D} materials for
  aerosolizable electronics.
\newblock {\em Nature Nanotechnology}, 2018.

\bibitem{korin12}
Netanel Korin et~al.
\newblock Shear-activated nanotherapeutics for drug targeting to obstructed
  blood vessels.
\newblock {\em Science}, 337:738--742, 2012.

\bibitem{kottapalli14}
A.~Kottapalli, M.~Asadnia, J.~Miao, and M.~Triantafyllou.
\newblock Touch at a distance sensing: lateral-line inspired {MEMS} flow
  sensors.
\newblock {\em Bioinspiration \& Biomimetics}, 9:046011, 2014.

\bibitem{li18}
S.~Li et~al.
\newblock A {DNA} nanorobot functions as a cancer therapeutic in response to a
  molecular trigger in vivo.
\newblock {\em Nature Biotechnology}, 36:258--264, 2018.

\bibitem{martel07a}
S.~Martel et~al.
\newblock Automatic navigation of an untethered device in the artery of a
  living animal using a conventional clinical magnetic resonance imaging
  system.
\newblock {\em Applied Physics Letters}, 90:114105, 2007.

\bibitem{martel07}
Sylvain Martel.
\newblock The coming invasion of the medical nanorobots.
\newblock {\em Nanotechnology Perceptions}, 3:165--173, 2007.

\bibitem{martel14}
Sylvain Martel et~al.
\newblock Computer {3D} controlled bacterial transports and aggregations of
  microbial adhered nano-components.
\newblock {\em J. of Micro-Bio Robotics}, 9:23--28, 2014.

\bibitem{merkle18}
Ralph~C. Merkle, Robert~A. {Freitas Jr.}, Tad Hogg, Thomas~E. Moore, Matthew~S.
  Moses, and James Ryley.
\newblock Mechanical computing systems using only links and rotary joints.
\newblock {\em ASME Journal on Mechanisms and Robotics}, 10:061006, 2018.

\bibitem{monroe09}
Don Monroe.
\newblock Micromedicine to the rescue.
\newblock {\em Communications of the ACM}, 52:13--15, June 2009.

\bibitem{murray26}
Cecil~D. Murray.
\newblock The physiological principle of minimum work: I. {The} vascular system
  and the cost of blood volume.
\newblock {\em Proc. of the Natl. Acad. of Sciences USA}, 12:207--214, 1926.

\bibitem{nagy09}
J.~A. Nagy, S-H. Chang, A.~M. Dvorak, and H.~F. Dvorak.
\newblock Why are tumour blood vessels abnormal and why is it important to
  know?
\newblock {\em British J. of Cancer}, 100:865--869, 2009.

\bibitem{painter06}
Page~R. Painter, Patrik Eden, and Hans-Uno Bengtsson.
\newblock Pulsatile blood flow, shear force, energy dissipation and {Murray's}
  law.
\newblock {\em Theoretical Biology and Medical Modeling}, 3:31, 2006.

\bibitem{park08}
Ji-Ho Park et~al.
\newblock Magnetic iron oxide nanoworms for tumor targeting and imaging.
\newblock {\em Advanced Materials}, 20:1630--1635, 2008.

\bibitem{pawlik81}
Gunter Pawlik, Angelika Rackl, and Richard~J. Bing.
\newblock Quantitative capillary topography and blood flow in the cerebral
  cortex of cats: an in vivo microscopic study.
\newblock {\em Brain Research}, 208:35--58, 1981.

\bibitem{purcell77}
Edward~M. Purcell.
\newblock Life at low {Reynolds} number.
\newblock {\em American Journal of Physics}, 45:3--11, 1977.

\bibitem{schulz09}
Mark~J. Schulz, Vesselin~N. Shanov, and Yeoheung Yun, editors.
\newblock {\em Nanomedicine Design of Particles, Sensors, Motors, Implants,
  Robots, and Devices}.
\newblock Engineering in Medicine and Biology. Artech House, Boston, 2009.

\bibitem{sershen00}
Scott Sershen, Sarah Westcott, N.~J. Halas, and J.~West.
\newblock Temperature-sensitive polymer-nanoshell composite for photothermally
  modulated drug delivery.
\newblock {\em J. of Biomedical Materials Research}, 51:293--298, 2000.

\bibitem{sherman81}
Thomas~F. Sherman.
\newblock On connecting large vessels to small: The meaning of {Murray's} law.
\newblock {\em The J. of General Physiology}, 78:431--453, 1981.

\bibitem{sichert09}
Andreas~B. Sichert, Robert Bamler, and J.~Leo van Hammen.
\newblock Hydrodynamic object recognition: When multipoles count.
\newblock {\em Physical Review Letters}, 102:058104, 2009.

\bibitem{sitti15}
M.~Sitti et~al.
\newblock Biomedical applications of untethered mobile milli/microrobots.
\newblock {\em Proceedings of the IEEE}, 103:205--224, 2015.

\bibitem{sochi15}
Taha Sochi.
\newblock Fluid flow at branching junctions.
\newblock {\em Intl. J. of Fluid Mechanics Research}, 42:59--81, 2015.

\bibitem{thompson92}
D'Arcy~Wentworth Thompson.
\newblock {\em On Growth and Form}.
\newblock Cambridge University Press, Cambridge, 1992.

\bibitem{thubagere17}
Anupama~J. Thubagere et~al.
\newblock A cargo-sorting {DNA} robot.
\newblock {\em Science}, 357:112, 2017.

\bibitem{vollmayr14}
Andreas~N. Vollmayr et~al.
\newblock Snookie: An autonomous underwater vehicle with artificial
  lateral-line system.
\newblock In H.~Bleckmann et~al., editors, {\em Flow Sensing in Air and Water},
  chapter~20, pages 521--562. Springer, 2014.

\bibitem{wu10}
Jing Wu, Daniel Day, and Min Gu.
\newblock Shear stress mapping in microfluidic devices by optical tweezers.
\newblock {\em Optics Express}, 18:7611--7616, 2010.

\bibitem{yang06a}
Yingchen Yang et~al.
\newblock Distant touch hydrodynamic imaging with an artificial lateral line.
\newblock {\em Proc. of the Natl. Acad. of Sciences USA}, 103:18891--18895,
  2006.

\end{thebibliography}
\end{document}